\documentclass{article}
\usepackage{ijcai22}

\usepackage{times}
\usepackage{soul}
\usepackage{url}
\usepackage[hidelinks]{hyperref}
\usepackage[utf8]{inputenc}
\usepackage[small]{caption}
\usepackage{graphicx}
\usepackage{amsmath}
\usepackage{amssymb}
\usepackage{booktabs}
\usepackage{algorithm}
\usepackage{algorithmic}
\newcommand{\our}{\mbox{RoSA}\space}
\newcommand{\eg}{\textit{e.g.}}
\newcommand{\ie}{\textit{i.e.}}

\title{RoSA: A Robust Self-Aligned Framework for Node-Node Graph Contrastive Learning}

\author{
Yun Zhu\footnotemark[1]
\and
Jianhao Guo\footnotemark[1]\and
Fei Wu\And
Siliang Tang\footnotemark[2]
\affiliations
Zhejiang University
\emails
\{zhuyun\_dcd,guojianhao,wufei,siliang\}@zju.edu.cn
}

\begin{document}

\maketitle

\renewcommand{\thefootnote}{\fnsymbol{footnote}}
\footnotetext[1]{Equal Contribution}
\footnotetext[2]{Corresponding Author}

\begin{abstract}
Graph contrastive learning has gained significant progress recently. However, existing works have rarely explored non-aligned node-node contrasting. In this paper, we propose a novel graph contrastive learning method named \our that focuses on utilizing non-aligned augmented views for node-level representation learning. First, we leverage the earth mover's distance to model the minimum effort to transform the distribution of one view to the other as our contrastive objective, which does not require alignment between views. Then we introduce adversarial training as an auxiliary method to increase sampling diversity and enhance the robustness of our model. Experimental results show that \our outperforms a series of graph contrastive learning frameworks on homophilous, non-homophilous and dynamic graphs, which validates the effectiveness of our work. To the best of our awareness, \our is the first work focuses on the non-aligned node-node graph contrastive learning problem. Our codes are available at: \href{https://github.com/ZhuYun97/RoSA}{\texttt{https://github.com/ZhuYun97/RoSA}}
\end{abstract}

\section{Introduction}
Graph representation learning, which aims to learn low dimension representations of nodes and edges for downstream tasks, has become a popular method when dealing with graph-domain data recently. Among all these methods, unsupervised graph contrastive learning has received considerable research attention. It combines the new research trend of graph neural network (GNN)~\cite{kipf2017semi} and contrastive self-supervised learning~\cite{oord2018representation,chen2020simple,grill2020bootstrap} methods, and has achieved promising results on many graph-based tasks~\cite{zhu2020deep,velickovic2019deep,you2020graph}.

\begin{figure}[htb]
    \centering
    \includegraphics[scale=0.4]{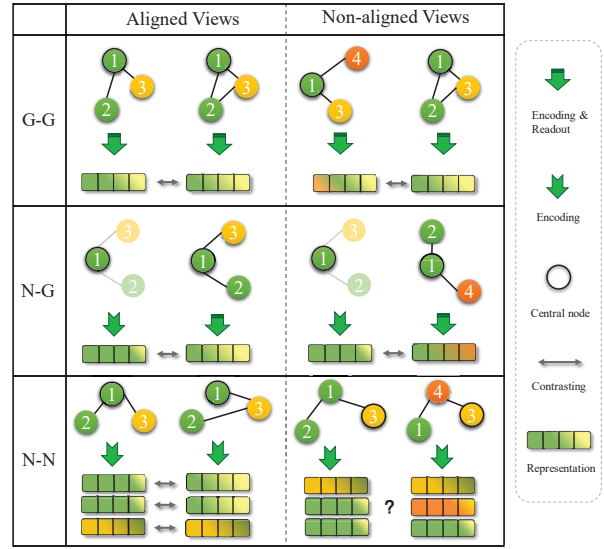}
    \caption{An illustration of different levels of contrasting methods, where G-G means graph-graph, N-G means node-graph and N-N means node-node contrasting level. We only show how a positive pair looks like, where the central node of subgraph is surrounded by a black circle. The number on nodes corresponds to their indices in the original full graph, and the color represents their labels.}
    \label{fig:contrast level}
\end{figure}

Contrastive learning aims to maximize the agreement between jointly sampled positive views and draw apart the distance between negative views, where in graph domain we refer augmented subgraph as a "view". 
Based on the scale of two contrasted views, graph contrasting learning can be classified as \emph{node-node}, \emph{node-graph}, and \emph{graph-graph} level~\cite{wu2021self}. 
From another perspective, a pair of contrasted views is recognized as \emph{aligned} or \emph{unaligned} depending on the difference of their node sets. Two aligned views must have identical node indices, except the structure and some features may differ, and two unaligned views can have different node sets. 
Figure~\ref{fig:contrast level} gives an illustrative overview according to this taxonomy.

\cite{zhu2021empirical} indicates that for node-level tasks such as node classification, applying node-node contrasting can obtain the best performance gain. However, previous work for node-node graph contrastive learning all contrast nodes in the aligned scenario which may hinder the flexibility and variability of sampled views and restrict the expressive power of contrastive learning. Moreover, there exist certain circumstances where aligned views are unavailable, for instance the dynamic graphs where the nodes may appear/disappear as time goes by, and the random walk sampling where the views are naturally non-aligned. Compared with aligned node-node contrasting, the non-aligned scenario is able to sample different nodes and their relations more freely, which will assist the model in learning more representative and robust features.

However, applying non-aligned node-node contrasting faces three challenges. First, how to design sub-sampling methods that can generate unaligned views while maintaining semantic consistency?
Second, how to contrast two non-aligned views even the number of nodes and correspondence between nodes are inconsistent?
Third, how to boost the performance meanwhile enhance the robustness of model for unsupervised graph contrastive learning? 
None of them have been satisfactorily answered by previous work.

To tackle the challenges discussed above, we propose RoSA: a \underline{\textbf{Ro}}bust \underline{\textbf{S}}elf-\underline{\textbf{A}}ligned framework for node-node graph contrastive learning. 
Firstly, we utilize random walk sampling to obtain augmented views for non-aligned node-node contrastive learning. Specifically, for a given graph, we sample a series of subgraphs based on a central node, and two different views of the same central node are treated as a positive pair, while views across different central nodes are selected as negative pairs. Note that even positive pairs are not necessarily aligned.
Secondly, inspired by the message passing mechanism of graph neural networks, the node representation can be interpreted as the result of distribution transformation of its neighboring nodes. Intuitively, for a pair of views, we leverage the earth mover's distance (EMD) to model the minimum effort to transform the distribution of one view to the other as our objective, which can implicitly align different views and capture the changes in their distributions.
Thirdly, we introduce unsupervised adversarial training that explicitly operates on node features to increase the diversity of samples and enhance the robustness of our model. 
To the best of our knowledge, this is the first work that fills the blank in non-aligned node-node graph contrastive learning. 

Our main contributions are summarized as follows:
\begin{itemize}
    \item We propose a robust self-aligned contrastive learning framework for node-node graph representation learning named RoSA. To the best of our knowledge, this is the first work dedicated to solving non-aligned node-node graph contrastive learning problems.
    \item To tackle the non-aligned problem, we introduce a novel graph-based optimal transport algorithm, \textit{g-EMD}, which does not require explicit node-node correspondence and can fully utilize graph topological and attributive information for non-aligned node-node contrasting. Moreover, to compensate for the possible information loss caused by non-aligned sub-sampling, we propose a non-trivial unsupervised graph adversarial training to improve the diversity of sub-sampling and strengthen the robustness of the model.
    \item Extensive experimental results on various graph settings achieve promising results and outperform several baseline methods by a large margin, which validates the effectiveness and generality of our method.
\end{itemize}

\section{Related Works}

\subsection{Self-Supervised Graph Representation Learning}
First appeared in the field of computer vision~\cite{oord2018representation,he2020momentum,grill2020bootstrap} and natural language processing~\cite{gao2021simcse}, self-supervised learning showed promising performance in various tasks and applying it to graph domain quickly became a research hot-spot. 
GraphCL~\cite{you2020graph} uses different augmentations and applies a readout function to obtain graph-graph level representations, then optimizes the InfoNCE loss, which can be mathematically proved to be the lower bound of mutual information.
Inspired by Deep InfoMax~\cite{hjelm2018learning}, DGI~\cite{velickovic2019deep} maximizes the mutual information between patch and graph representations, which is node-graph level contrasting. 
Recently, node-node level methods like GMI~\cite{peng2020graph}, GRACE~\cite{zhu2020deep}, GCA~\cite{zhu2021graph} and BGRL~\cite{thakoor2021bootstrapped} show superior performance on node classification task. Unlike DGI, GMI removes the readout function and maximizes the MI between inputs and outputs of the encoder at the node-node level. With graph augmentation methods, GRACE focuses on contrasting aligned views using different nodes as the negative pairs, and the same nodes from different views are regarded as positive pairs, where each positive pair should be aligned first. GCA is similar to GRACE but with adaptive data augmentation. BGRL is a negative-sample-free method which borrows the idea from BGRL~\cite{grill2020bootstrap}.

Previous works that involve graph level contrasting, usually have a readout function to obtain whole graph representation, which are naturally aligned, but when it comes to node-node level contrasting, they always explicitly align nodes for positive pairs. 
The work of non-aligned node-node graph contrastive learning has not yet been explored.

\subsection{Adversarial Training}

Adversarial training (AT) has been found useful to improve the model's robustness. 
AT is a min-max training process, which aims to maintain the consistency of the model's output before and after adding adversarial perturbations.
Previous works solve the adversarial perturbations from many different perspectives. \cite{goodfellow2014explaining} gives a linear approximation of the perturbation under L2 norm (\ie Fast Gradient method).
Projected Gradient Descent method~\cite{madry2018towards} tries to obtain a more precise perturbation in an iterative manner, but it takes more time, \cite{shafahi2019adversarial,zhu2020freelb} provide more efficient methods. Lately, \cite{kong2020flag} adopts these methods into the graph domain in a supervised manner. However, unsupervised adversarial training for graphs is still unexplored. In this paper, we adopt AT into our contrastive method to improve the robustness of the model in an unsupervised manner.

\begin{figure*}[htbp]
    \centering
    \includegraphics[scale=0.55]{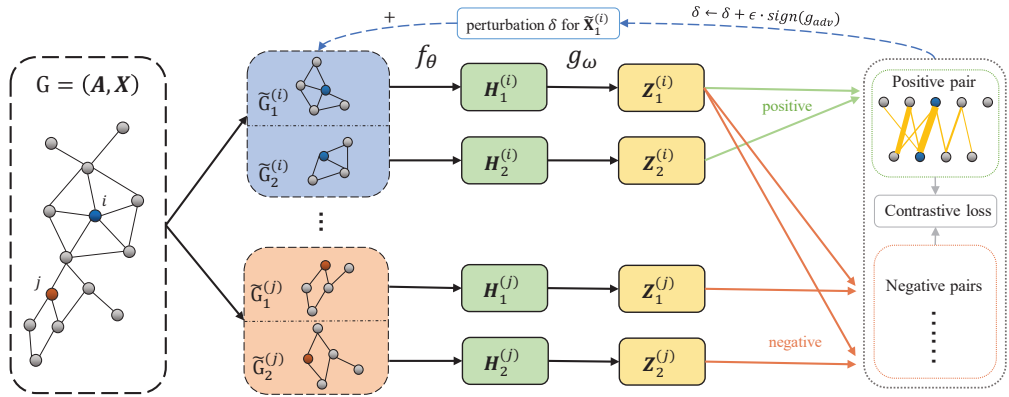}
    \caption{The overview of our proposed method: RoSA. The input is a series of subgraphs sampled from a full graph, where different random walk views from the same central node are recognized as positive pairs and views from different central nodes are treated as negative pairs. Then the subgraphs are fed into the encoder and projector to obtain node embeddings for contrasting. The self-aligned EMD-based contrastive loss will maximize the mutual information (MI) between positive pairs and minimize MI between negative pairs, guiding the model to learn rich representations. Besides, introducing adversarial training into this workflow enhances the robustness of the model.}
    \label{figure:overview}
\end{figure*}
\section{Method}

In this chapter, we will introduce the framework of RoSA. Figure \ref{figure:overview} gives an overview of RoSA.

\subsection{Preliminaries}
Given a graph $\mathcal{G} = (\mathcal{V}, \mathcal{E})$, where $\mathcal{V}$ is the set of $N$ nodes and $\mathcal{E}$ is the set of $M$ edges. Also use $\mathcal{G}=(\mathbf{X}, \mathbf{A})$ to represent graph features, where $\mathbf{X}=\left\{\mathbf{x}_{1}, \mathbf{x}_{2}, \ldots, \mathbf{x}_{N}\right\} \in R^{N \times d}\ $ represents node feature matrix, each node's feature dimension is $d$ and can be formulated as $\mathbf{x}_{i} \in \mathbb{R}^{d}, \ \mathbf{A} \in \mathbb{R}^{N \times N}$ represents the graph adjacency matrix, where $\mathbf{A}_{i,j}=1$ if an edge exists between node $i$ and $j$, else $\mathbf{A}_{i,j}=0$. 
For subgraph sampling, each node $i$ will be treated as central node to get subgraph $\mathcal{G}^{(i)}$. An augmented view of subgraph $\mathcal{G}^{(i)}$ is represented as $\tilde{\mathcal{G}}^{(i)}_{k}$ where subscript $k$ denotes the $k$-th augmented view.

\subsection{Non-Aligned Node-Node Level Sub-Sampling}
It has been proven that well-designed data augmentation plays a vital role in boosting the performance of contrastive learning~\cite{you2020graph}. However, different from the CV and NLP domain, where data is organized in a Euclidean fashion, graph data augmentation methods need to be redesigned and carefully selected.

It is worth noting that in this work, for a positive pair, we need to get different sets of nodes while preserving the consistency of their semantic meanings. Based on such a premise, we propose to utilize random walk with restart sampling as an augmentation method that selects nodes randomly and generates unaligned views.
Specifically, random walk sampling starts from the central node $v$ and generates a random path with a given step size $s$. Besides, at each step the walk returns to central node $v$ with a restart probability $\alpha$.
The step size $s$ should not be too large because we want to capture the local structure of the central node. Lastly, edge dropping and feature masking \cite{zhu2020deep} are applied on subgraphs.

\subsection{g-EMD: A Self-aligned Contrastive Objective}
After obtaining two unaligned augmented views, we define a contrastive objective that measures the agreement of two different views. Prior arts mostly use cosine similarity as a metric to evaluate how far two feature vectors drift apart. While under our setting, two views may have different and unaligned nodes, where a simple cosine similarity loses its availability. Hence we propose to leverage the earth mover's distance (EMD) as our similarity measure. 

EMD~\cite{rubner2000earth,zhang2020deepemd,liu2020self} is the measure of the distance between two discrete distributions, it can be interpreted as the minimum cost to move one pile of dirt to the other. Although prior work has introduced EMD to the CV domain, the adaptation in the graph domain has not been explored yet. Moreover, according to the characteristics of graph data, we also take topology distance into consideration while computing the cost matrix. Through a non-trivial solution, we combine the vanilla cost matrix and topology distance to obtain a rectified cost matrix which makes the cost related to the node similarity and the distance in the graph topology.

The calculation of \textit{g-EMD} can be formulated as a linear optimization problem. In our case, the two augmented views have feature maps $\mathbf{X} \in \mathbb{R}^{M \times d}$ and $\mathbf{Y} \in \mathbb{R}^{N \times d}$ respectively, the goal is to measure the distance to transform $\mathbf{X}$ to $\mathbf{Y}$. Suppose for each node $\boldsymbol{x}_i \in \mathbb{R}^{d}$, it has $\boldsymbol{t}_i$ units to transport, and node $\boldsymbol{y}_j \in \mathbb{R}^{d}$ has $\boldsymbol{r}_j$ units to receive. For a given pair of nodes $\boldsymbol{x}_i$ and $\boldsymbol{y}_j$, the cost of transportation per unit is $\mathbf{D}_{ij}$, and the amount of transportation is $\mathbf{\Gamma}_{ij}$. With above notations, we can define the linear optimization problem as follows:
\begin{align}
    \min\limits_{\mathbf{\Gamma}}   &\sum\limits^{M}_{i}\sum\limits^{N}_{j}\mathbf{D}_{ij}\mathbf{\Gamma}_{ij},\\
    s.t. &\mathbf{\Gamma}_{ij} \geq 0,i=1, 2, ..., M, j=1, 2, ..., N\notag \\
        &\sum\limits^{M}_{i}\mathbf{\Gamma}_{ij} = r_j, j=1, 2, ..., N \notag \\
        &\sum\limits^{N}_{j}\mathbf{\Gamma}_{ij} = t_i, i=1, 2, ..., M \notag
\end{align}
where $\textbf{t} \in \mathbb{R}^{M}$ and $\textbf{r} \in \mathbb{R}^{N}$ are marginal weights for $\mathbf{\Gamma}$ respectively.

The set of all possible transportation matrices $\Gamma$ can be defined as
\begin{equation}
    \Pi(\mathbf{t},\mathbf{r}) = \{\mathbf{\Gamma} \in \mathbb{R}^{M \times N}|\mathbf{\Gamma} \mathbf{1}_M = \mathbf{t},\mathbf{\Gamma}^T \mathbf{1}_N=\mathbf{r}\},
\end{equation}
where $\mathbf{1}$ is all-one vector with corresponding size, and $\Pi(\mathbf{t},\mathbf{r})$ is the set of all possible distributions whose marginal weights are $\mathbf{t}$ and $\mathbf{r}$.

And the cost to transfer $\boldsymbol{x}_i$ to $\boldsymbol{y}_j$ is defined as
\begin{equation}
    \mathbf{D}_{ij} = 1 - \frac{{\boldsymbol{x}^T_i}{\boldsymbol{y}_j}}{||\boldsymbol{x}_i||||\boldsymbol{y}_j||},
\end{equation}
which indicates that nodes with similar representations prefer to generate fewer matching cost between each other. In addition to directly using node representations dissimilarity matrix as a distance matrix, we also take the topology distance $\Psi \in \mathbb{R}^{M \times N}$(the smallest hop count between each pair of nodes) into consideration. Nodes are close in topology structure which indicates they may contain similar semantic information. How to combine the representation dissimilarity matrix and topology distance is not a trivial problem. In order not to adjust the original cost matrix $\mathbf{D}$ drastically, we adopt sigmoid function $S$ with temperature on topology distance to get re-scale factors $\mathbf{S} \in [0.5, 1]^{M \times N}$:
\begin{equation}
\mathbf{S}_{i,j}=S(\Psi_{i,j})=\frac{1}{1+e^{-\Psi_{i,j}/\tau}},
\end{equation}
where $\tau \geq 1$ is the temperature factor to control the rate of curve change. We set $\tau$ as 2 empirically, and leave the choice of different of re-scale function and the tuning of different temperature factors in future work. With the re-scale factors $\mathbf{S}$, we can update the cost matrix by
\begin{equation}
    \mathbf{D} = \mathbf{D} \circ \mathbf{S},
\end{equation}
where $\circ$ is Hadamard product. In this way, we combine both topology distance and node representation dissimilarity matrix into distance matrix. 

As $\mathbf{D}$ is fixed according to distributions $\mathbf{X}$, $\mathbf{Y}$ and topology distance, to get g-EMD we need to find the optimal $\tilde{\mathbf{\Gamma}}$. To solve the optimal $\tilde{\mathbf{\Gamma}}$, we utilize \emph{Sinkhorn Algorithm}~\cite{cuturi2013sinkhorn} by introducing a regularization term:
\begin{equation}
    \mathrm{g\text{-}EMD}(\mathbf{X},\mathbf{Y},\mathbf{S}) = \inf_{\mathbf{\Gamma} \in \Pi} \langle \mathbf{\Gamma}, \mathbf{D}\rangle_\mathbf{\mathrm{F}} + \underbrace{\frac{1}{\lambda}\mathbf{\Gamma}(\log {\mathbf{\Gamma}}-1)}_{\text{regularization term}},
\end{equation}
where $\langle , \rangle_\mathrm{F}$ denotes Frobenius inner product, and $\lambda$ is a hyperparameter that controls the strength of regularization. With this regularization, the optimal $\tilde{\Gamma}$ can be approximated as:
\begin{equation}
    \tilde{\mathbf{\Gamma}} = diag(\mathbf{v})\mathbf{P} diag(\mathbf{u}),
\label{eq7}
\end{equation}
where $\mathbf{P}=e^{-\lambda \mathbf{D}}$, and  $\mathbf{v}$, $\mathbf{u}$ are two coefficient vectors whose values can be iteratively updated as
\begin{align}
    & \boldsymbol{v}^{t+1}_i = \frac{\boldsymbol{t}_i}{\sum^N_{j=1}{\mathbf{P}_{ij}\boldsymbol{u}^t_j}}, \\
    & \boldsymbol{u}^{t+1}_j = \frac{\boldsymbol{r}_j}{\sum^M_{i=1}{\mathbf{P}_{ij}\boldsymbol{v}^{t+1}_i}}. \notag
\end{align}

Then the question lies in how to get marginal weights $\textbf{t}$ and $\textbf{r}$. 
The weight represents a node's contribution in comparison of two views, where a node should have larger weight if its semantic meaning is close to the other view. Based on this hypothesis, we define the node weight as dot product between its feature and the mean pooling feature from the other set:
\begin{align}
    & \boldsymbol{t}_i = \max\{{\boldsymbol{x}^T_i \cdot \frac{\sum^N_{j=1}{\boldsymbol{y}_j}}{N},0}\}, \\
    & \boldsymbol{r}_j = \max\{{\boldsymbol{y}^T_j \cdot \frac{\sum^M_{i=1}{\boldsymbol{x}_i}}{M},0}\}. \notag
\end{align}
where $max$ is to make sure all weights are non-negative, and then both views will be normalized to ensure having the same amount of features to transport. 

With optimal transportation amount $\tilde{\mathbf{\Gamma}}$, we obtain:
\begin{equation}
    \mathrm{g\text{-}EMD}(\mathbf{X},\mathbf{Y},\mathbf{S}) = \langle \tilde{\mathbf{\Gamma}}, \mathbf{D}\rangle_\mathrm{F}.
\end{equation}

Now we can leverage EMD as the distance measure to contrastive loss objective. For any central node $\boldsymbol{v}_i$ and its augmented graph views $(\tilde{\mathcal{G}}^{(i)}_{1}, \tilde{\mathcal{G}}^{(i)}_{2})$, an encoder $f_\theta$ (e.g. GNN) is applied to get embeddings $\mathbf{H}^{(i)}_1$ and $ \mathbf{H}^{(i)}_2$ respectively, then a linear projector $g_\omega$ is applied on top of that to get $\mathbf{Z}^{(i)}_1$ and $ \mathbf{Z}^{(i)}_2$ to improve generality for downstream tasks as indicated in~\cite{chen2020simple}. Formally, we define the EMD-based contrastive loss for node $\boldsymbol{v}_i$ as
\begin{multline}
    \ell(\mathbf{Z}^{(i)}_1, \mathbf{Z}^{(i)}_2) = \\
    -\log (\frac{e^{\textnormal{s}(\mathbf{Z}^{(i)}_1, \mathbf{Z}^{(i)}_2))/ \tau}}{\sum_{k=1}^{N}e^{\textnormal{s}(\mathbf{Z}^{(i)}_1, \mathbf{Z}^{(k)}_2))/ \tau} +  \sum_{k=1}^{N} \mathbf{1}_{[k \neq i]} e^{\textnormal{s}(\mathbf{Z}^{(i)}_1, \mathbf{Z}^{(k)}_1))/ \tau} }),
\end{multline}
where $\textnormal{s}(\boldsymbol{x}, \boldsymbol{y})$ is a function that calculates the similarity between $\boldsymbol{x}$ and $\boldsymbol{y}$, here we use $1-\rm{EMD}(\boldsymbol{x},\boldsymbol{y})$ to replace $\textnormal{s}(\boldsymbol{x}, \boldsymbol{y})$; $\mathbf{1}$ is an indicator function which returns $1$ if $i \neq k$ otherwise returns $0$; and $\tau$ is temperature parameter. Adding all nodes in $\mathcal{N}$, the overall contrastive loss is given by: 
\begin{equation}
\mathcal{J}=\frac{1}{2 N} \sum_{i=1}^{N}\left[\ell\left(\mathbf{Z}^{(i)}_1, \mathbf{Z}^{(i)}_2\right)+\ell\left(\mathbf{Z}^{(i)}_2, \mathbf{Z}^{(i)}_1\right)\right].
\label{loss}
\end{equation}

We summarize our proposed algorithm for non-aligned node-node contrastive learning in Appendix A.

\subsection{Unsupervised Adversarial Training}
Adversarial training can be considered as an augmentation technique which aims to improve the model's robustness. \cite{kong2020flag} has empirically proven that graph adversarial augmentation on feature space can boost the performance of GNN under a supervised manner. Such a method can be modified for graph contrastive learning as

\begin{equation}
\underset{\theta, \omega}{\operatorname{min}} \mathbb{E}_{(\mathbf{X}_1^{(i)}, \mathbf{X}_2^{(i)}) \sim \mathbb{D}}\left[\frac{1}{M} \sum_{t=0}^{M-1} \max _{\boldsymbol{\delta}_{t} \in \mathcal{I}_{t}} \mathcal{J}\left(\mathbf{X}_1^{(i)}+\boldsymbol{\delta}_{t},\mathbf{X}_2^{(i)}\right)\right],
\end{equation}
where $\theta, \omega$ are the parameters of encoder and projector, $\mathbb{D}$ is data distribution, $\mathcal{I}_{t}=\mathcal{B}_{\mathbf{X}+\boldsymbol{\delta}_{0}}(\alpha t) \cap \mathcal{B}_{\mathbf{X}}(\epsilon)$ where $\epsilon$ is the perturbation budget. For efficiency, the inner loop runs $M$ times, the gradient of $\delta, \theta_{t-1}\ \text{and}\ \omega_{t-1}$ will be accumulated in each time, and the accumulated gradients will be used for updating $\theta_{t-1}\ \text{and}\ \omega_{t-1}$ during outer update. Equipped with such adversarial augmentation, we complete a more robust self-aligned task. The energy is hopefully transferred between nodes belonging to different categories during max-process, and min-process will remedy such a bad situation to make the alignment more robust. In this way, the adversarial augmentation increases the diversity of samples and improves the robustness of the model.

\section{Experiments}
We conduct extensive experiments on ten public benchmark datasets to evaluate the effectiveness of RoSA. We use \our to learn node representations in an unsupervised manner and assess their quality by a linear classifier trained on top of that. Some more detailed information about datasets and experimental setup can be found in Appendix B, C.

\subsection{Datasets}
We conduct experiments on ten public benchmark datasets that include four homophilous datasets (Cora, Citeseer, Pubmed and DBLP), three non-homophilous datasets (Cornell, Wisconsin and Texas), two large-scale inductive datasets (Flickr and Reddit) and one dynamic graph dataset (CIAW) to evaluate the effectiveness of RoSA. Details of datasets can be found in Appendix B.

\subsection{Experimental Setup}

\paragraph{Models} 
For small-scale datasets, we apply a two-layer GCN as our encoder $f_\theta$  and for the large-scale datasets (Flickr and Reddit), we adopt a three-layer GraphSAGE-GCN~\cite{hamilton2017inductive} with residual connections as the encoder following DGI~\cite{velickovic2019deep} and GRACE~\cite{zhu2020deep}. The formulas of encoders can be found in Appendix C.
Specifically, similar to \cite{chen2020simple}, a projection head which comprises a two-layer non-linear MLP with BN is added on top of the encoder. Detailed hyperparameter settings are in Appendix C.

\paragraph{Baselines}
We compare \our with two node-graph constrasting methods DGI~\cite{velickovic2019deep}, SUBG-CON~\cite{jiao2020subgraph}), and four node-node methods GMI~\cite{peng2020graph}, GRACE~\cite{zhu2020deep}, GCA~\cite{zhu2021graph} and BGRL~\cite{thakoor2021bootstrapped}.
\subsection{Results and Analysis}

\begin{table}[htb]
\scalebox{0.8}{
\begin{tabular}{llllll}
\toprule
Method                     & Level & Cora     & Citeseer & Pubmed   & DBLP    \\ \hline
Raw Features               & -     & 64.8     & 64.6     & 84.8     & 71.6    \\
DeepWalk                   & -     & 67.2     & 43.2     & 65.3     & 75.9    \\
GCN                        & -     & 82.8     & 72.0     & 84.9     & 82.7    \\ \hline
% GAE                        & -     & 76.9     & 60.6     & 82.9     & 81.2    \\
% VGAE                       & -     & 78.9     & 61.2     & 83.0     & 81.7    \\
DGI                        & N-G   & 82.6±0.4 & 68.8±0.7 & 86.0±0.1 & 83.2±0.1\\
$\text{SUBG-CON}^{\star}$  & N-G   & 82.6±0.9 & 69.2±1.3 & 84.3±0.3 & 83.8±0.3\\
GMI                        & N-N   & 82.9±1.1 & 70.4±0.6 & 84.8±0.4 & 84.1±0.2\\
GRACE                      & N-N   & 83.3±0.4 & 72.1±0.5 & 86.7±0.1 & 84.2±0.1\\ 
GCA                        & N-N   & 83.8±0.8 & 72.2±0.7 & 86.9±0.2 & 84.3±0.2\\
BGRL                       & N-N   & 83.8±1.6 & 72.3±0.9 & 86.0±0.3 & 84.1±0.2\\ \hline 
RoSA                       & N-N   & \textbf{84.5±0.8} & \textbf{73.4±0.5}   & \textbf{87.1±0.2} & \textbf{85.0±0.2}\\

\bottomrule
\end{tabular}}
\caption{Summary of classification accuracy of node classification tasks on homophilous graphs.
The second column represents the contrasting mode of methods, N-G stands for node-graph level, and N-N stands for node-node level. For a fair comparison, in $\text{SUBG-CON}^{\star}$ we replace the original encoder with the encoder used in our paper and apply the same evaluation protocol as ours.}
\label{table:main}
\end{table}

\paragraph{Results for homophilous datasets}
Table \ref{table:main} shows the node classification results on four homophilous datasets, some of the reported statistics are borrowed from~\cite{zhu2020deep}. Experiment results show that N-N methods surpass N-G on node classification tasks. And \our is superior to all baselines and achieves state-of-the-art performance, and even surpasses the supervised method (GCN), which proves the effectiveness of leveraging EMD-based contrastive loss and adversarial training in non-aligned node-node scenarios. Different from other node-node methods that train on full graphs, our method is trained on various non-aligned subgraphs, which brings more flexibility but also non-alignment challenge. \our learns more information from the challenging pretext task. The visualization of cost matrix and transportation matrix in EMD during training is in Appendix E.

\paragraph{Results for non-homophilous dataset}

% \begin{table}[htb]
% \begin{tabular}{lllll}
% \toprule
% Methods                & Cornell  & Wisconsin & Texas    \\ 
% \midrule
% DGI                  & 56.3±4.7       & 50.9±5.5  & 56.9±6.3 \\
% $\text{SUBG-CON}$& 54.1±6.7 & 48.3±4.8  & 56.9±6.9 \\
% GRACE                  & 58.2±4.1 & 54.3±7.1  & 58.9±4.7 \\ \hline
% \our      & \textbf{59.3±3.6} & \textbf{55.1±4.7}  & \textbf{60.3±4.5} \\ 
% \bottomrule
% \end{tabular}
% \caption{Non-homophilous node classification using GCN.}
% \label{table:gnn}
% \end{table}

\begin{table}[htb]
\scalebox{0.7}{
\begin{tabular}{llll|lll}
\toprule
Methods            & Cornell    & Wiscons. & Texas    & Cornell    & Wiscons. & Texas   \\
\midrule
DGI                & 56.3±4.7   & 50.9±5.5  & 56.9±6.3 & 58.1±4.1   & 52.1±6.3  & 57.8±5.2 \\
$\text{SUBG-CON}$  & 54.1±6.7   & 48.3±4.8  & 56.9±6.9 & 58.7±6.8   & 59.0±7.8  & 61.1±7.3 \\
GMI                & 58.1±4.0   & 52.9±4.2  & 57.8±5.9 & 69.6±5.3   & 70.8±5.2  & 69.6±5.3 \\
GRACE              & 58.2±4.1   & 54.3±7.1  & 58.9±4.7 & 72.3±5.3   & 74.1±5.5  & 69.8±7.2 \\ \hline
% $\text{GRACE}^{\dag}$     & 68.0±5.9 & 71.6±5.4   &67.4±6.3\\ 
RoSA    & \textbf{59.3±3.6} & \textbf{55.1±4.7}  & \textbf{60.3±4.5} & \textbf{74.3±6.2}   & \textbf{77.1±4.3}  & \textbf{71.1±6.6} \\
\bottomrule
\end{tabular}}
\caption{Non-homophilous node classification using GCN (left) and MLP (right).}
\label{table:mlp}
\end{table}

Previous works have shown that GCN performs poorly on non-homophilous graphs~\cite{pei2020geom,zhu2020homophily}, because there are a lot of high-frequency signals on such graphs, and GCN is essentially a low-pass filter, where a lot of useful information will be filtered out. Since the design of the encoder is not the focus of our work, we use both GCN and MLP as our encoders in this part.

We compare the performance of our model with DGI, SUBG-CON, GMI, GRACE using either GCN or MLP as encoder, see Table \ref{table:mlp}. From the statistics, we can summarize three major conclusions:
Firstly, the overall performance of SUBG-CON and DGI lags behind the others. This is because SUBG-CON and DGI are the node-graph level contrasting methods that maximize the mutual information between central node representation and its contextual subgraph representation, and under the non-homophilous circumstance, the contextual graph representation gathers highly variant features from different kinds of nodes, which renders wrong and meaningless signals.

Secondly, with the same method, the MLP version performs significantly better than its GCN counterpart, which confirms the statement that MLP is more suitable for non-homophilous graphs. Furthermore, we can observe that the performance gap between node-global and node-node methods widens when using MLP as the encoder. We suspect such a phenomenon is caused because the GCN encoder loses a large amount of information under a non-homophilous setting and makes the effort of other modules in vain.

Thirdly and most importantly, \our outperforms other benchmarks on all three datasets, no matter the choice of the encoder, which validates the effectiveness of \our for non-homophilous graphs. We speculate that \our will tighten the distance of nodes of the same class.

\paragraph{Result for inductive learning on large-scale datasets}
The experiments conducted above are all under the transductive setting. In this part, the experiments are under the inductive setting where tests are conducted on unseen or untrained nodes. The micro-averaged F1 score is used for both of these two datasets. 
The results are shown in Table \ref{table:inductive}, we can see that \our works well on large-scale graphs under inductive setting and reaches state-of-the-art performance. 
An explanation is DGI, GMI and GRACE can not directly work on full graphs, they use the sampling strategy proposed by \cite{hamilton2017inductive} in their original work. However, we adopt subsampling (random walk) as our augmentation technique which means our method can seamlessly work on these large graphs. Furthermore, our pretext task is designed for such subsampling which is more suitable for large graphs.

\begin{table}[htb]
\centering
\begin{tabular}{cccc}
\toprule
Methods                    & Flickr   & Reddit                        \\
\midrule
Raw features        &  20.3    & 58.5                         \\
DeepWalk            &  27.9    & 32.4                           \\ 
FastGCN             & 48.1±0.5 & 89.5±1.2                   \\
GraphSAGE           & 50.1±1.3 & 92.1±1.1                   \\\hline
% DeepWalk + features &          & 69.1      & —                            \\ \hline
Unsup-GraphSAGE     &  36.5    & 90.8                        \\

DGI                 & 42.9±0.1 & 94.0±0.1                  \\
GMI                 & 44.5±0.2 & 95.0±0.0                    \\
GRACE               & 48.0±0.1 & 94.2±0.0                     \\ \hline
RoSA                & \textbf{51.2±0.1} & \textbf{95.2±0.0}     \\

\bottomrule
\end{tabular}
\caption{Result for inductive learning on large-scale datasets.}
\label{table:inductive}
\end{table}

\paragraph{Results for dynamic graphs dataset}
In addition, we also test our method on dynamic graphs.
For the contrastive task, we consider the adjacent snapshots as positive views because the evolution process is generally "smooth", and the snapshots far away from the anchor are considered as negative views. 
In CIAW, each snapshot maintains all nodes appeared in the timeline, however, in real-world scenarios, the addition or deletion of nodes happens as time goes by.
So in CIAW*, we remove isolated nodes in each snapshot to emulate such a situation. 
Note that GRACE can not work on CIAW* because CIAW* creates a non-aligned situation, while GRACE is inherently an aligned method.
From the statistics in Table \ref{table:dynamic}, \our surpasses other competitors and can work well in both situations. 
Currently, we simply use static GNN encoder with discrete-time paradigms  which can be replaced with temporal GNN encoders, and we will leave it for future work.
\begin{table}[htb]
\centering
\begin{tabular}{ccc}
\toprule
          & CIAW     & CIAW*    \\ 
\midrule
GraphSAGE & 64.0±8.5 & 69.7±10.1 \\ 
GRACE     & 65.3±7.9 & -        \\
RoSA      & \textbf{67.6±7.0} & \textbf{73.2±9.3}\\
\bottomrule

\end{tabular}
\label{ciaw}
\caption{Node classification using GraphSAGE on dynamic graphs.}
\label{table:dynamic}
\end{table}

\subsection{Ablation Study}
\begin{figure}[htb]
    \centering
    \includegraphics[scale=0.4]{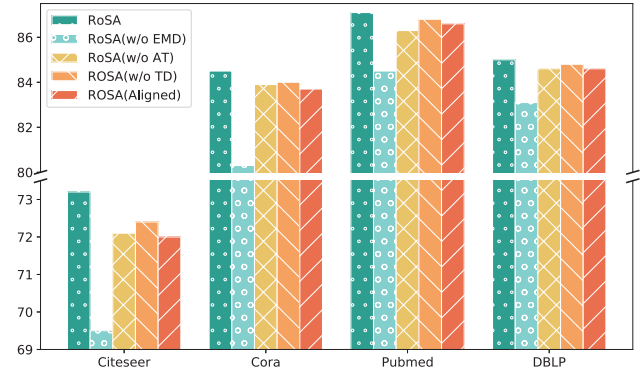}
    \caption{Abalation study on RoSA}
    \label{fig:ablation}
\end{figure}
To prove the effectiveness of the design of RoSA, we conduct ablation experiments masking different components under the same hyperparameters. First we replace the EMD-based InfoNCE loss with a regular cosine similarity metric, represented as \our w/o EMD (In order to make it computable under such situation, we restrict the same amount of nodes for contrasted views). Second we use the vanilla cost matrix for EMD, named as RoSA w/o TD. Then we remove the adversarial training process, denoted as \our w/o AT. Finally, we adopt aligned views contrasting instead of the original non-aligned random walking, named as \our Aligned. For a fair comparison, we keep other hyperparameters and the training scheme same. The results is summarized in Figure \ref{fig:ablation}. As we can see, the performance degrades without either EMD, adversarial training or rectified cost matrix, which indicates the effectiveness of the corresponding components. Furthermore, compared to aligned views, the model achieves comparable or even better results under the non-aligned condition, which demonstrates that our model, to a certain degree, solves the non-aligned graph contrasting problem. The experiments of sensitivity analysis are in Appendix D.

\section{Conclusion}
In this paper, we propose a robust self-aligned framework for node-node graph contrastive learning, where we design and utilize the graph-based earth mover's distance (\textit{g-EMD}) as a similarity measure in the contrastive loss to avoid explicit alignment between contrasted views. Then we introduce unsupervised adversarial training into graph domain to further improve the robustness of the model. Extensive experiment results on homophilous, non-homophilous and dynamic graphs datasets demonstrate that our model can effectively be applied to non-aligned situations and outperform other competitors. Moreover, in this work we adopt simple random walk with restart as the subsampling technique, and RoSA may achieve better performance if equipped with more powerful sampling methods in future work.

% \input{sections/appendix}

%% The file named.bst is a bibliography style file for BibTeX 0.99c
\bibliographystyle{named}
\newpage
\bibliography{ref}
\newpage
\quad
\newpage
\newpage

\subsection*{A. Algorithm}
The steps of the procedure of our method are summarised as:
\begin{algorithm}[htb]
    \textbf{Input}: Sampler function $\mathcal{T}(\mathcal{G}, \text{idx})$, ascent steps $M$, ascent step size $\alpha$, additional augmentations $\tau_{\alpha}, \tau_{\beta}$, encoder $f_\theta$, projector $g_\omega$, and training graph $\mathcal{G}=\{\mathbf{A}, \mathbf{X}\}$ \\
    % \textbf{Output}:  \TODO{}\\
    \begin{algorithmic}[1] %[1] enables line numbers
        \WHILE{not converge}
        \FOR{$i = 1$ to $N$}
        \STATE Sample two context subgraphs $\mathcal{G}^{(i)}_1 = \mathcal{T}(\mathcal{G}, i)$, $\mathcal{G}^{(i)}_2 = \mathcal{T}(\mathcal{G}, i)$ for each central node
        \ENDFOR
        \STATE $\mathbf{S} = \{( \mathcal{G}^{(i)}_1, \mathcal{G}^{(i)}_2)\}, i=1...N$
        \STATE sampled batch $\mathcal{B}=\{( \mathcal{G}^{(k)}_1, \mathcal{G}^{(k)}_2)\} \in \mathbf{S}$
        \STATE $\mathcal{B}_1 = \{\mathcal{G}^{(k)}_1\}, \mathcal{B}_2 = \{\mathcal{G}^{(k)}_2\}$
        \STATE $\{(\tilde{\mathbf{X}}^{(k)}_1, \tilde{\mathbf{A}}^{(k)}_1)\} =\tilde{\mathcal{B}}_1 = \tau_{\alpha}(\mathcal{B}_1)$
        \STATE $\{(\tilde{\mathbf{X}}^{(k)}_2, \tilde{\mathbf{A}}^{(k)}_2)\} =\tilde{\mathcal{B}}_2 = \tau_{\beta}(\mathcal{B}_2)$
        
        \STATE $\delta_{0} \leftarrow U(-\alpha, \alpha)$
        \STATE $g_{0} \leftarrow 0$
        \FOR{$\mathrm{t}=1 \ldots M$} 
        \STATE $\mathbf{Z}_1=g_{\boldsymbol{\omega}} \circ f_{\boldsymbol{\theta}}\left(\mathbf{\tilde{X}}_{1}+\boldsymbol{\delta}_{t-1} , \tilde{\mathbf{A}}_1\right)=g_\omega \circ f_\theta(\mathcal{\tilde{B}}_1+\boldsymbol{\delta}_{t-1})$
        \STATE $\mathbf{Z}_2 = g_\omega \circ f_\theta(\mathcal{\tilde{B}}_2)$
        \STATE $\boldsymbol{g}_{t} \leftarrow \boldsymbol{g}_{t-1}+\frac{1}{M} \cdot \nabla_{\boldsymbol{\theta}, \boldsymbol{\omega}} \ell\left(\mathbf{Z}_{1}, \mathbf{Z}_{2}\right)$
        \STATE $\boldsymbol{g}_{\boldsymbol{\delta}} \leftarrow \nabla_{\boldsymbol{\delta}} \ell\left(\mathbf{Z}_{1}, \mathbf{Z}_{2}\right)$
        \STATE $\delta_{t} \leftarrow \delta_{t-1}+\alpha \cdot \boldsymbol{g}_{\delta} /\left\|\boldsymbol{g}_{\boldsymbol{\delta}}\right\|_{F}$
        \ENDFOR
        \STATE $\boldsymbol{\theta} \leftarrow \boldsymbol{\theta}-\tau \cdot \boldsymbol{g}_{M, \theta}$
        \STATE $\boldsymbol{\omega} \leftarrow \boldsymbol{\omega}-\tau \cdot \boldsymbol{g}_{M, \omega}$
        \ENDWHILE
        % \STATE \textbf{return} \TODO{}
    \end{algorithmic}
    \caption{Algorithm for a robust self-aligned framework for node-node graph contrastive learning.}
    \label{alg1}
\end{algorithm}

\begin{table*}[tb]
\begin{tabular}{llllllllllllll}
\hline
          & \begin{tabular}[c]{@{}l@{}}Hidden\\ size\end{tabular} & \begin{tabular}[c]{@{}l@{}}Batch\\ size\end{tabular} & \begin{tabular}[c]{@{}l@{}}Learning\\ rate\end{tabular} & \begin{tabular}[c]{@{}l@{}}Weight\\ decay\end{tabular} & \begin{tabular}[c]{@{}l@{}}Walk\\ length\end{tabular} & Epochs & Patience & Optimizer & $\tau$ & $p_{e,1}$ & $p_{e,2}$ & $p_{f1}$ & $p_{f2}$ \\ \hline
Cora      & 128 & 128 & 1e-2 & 5e-4 & 10 & 500 & - & SGD   & 0.4  & 0.2 & 0.2 & 0.3 & 0.3 \\
Citeseer  & 256 & 128 & 1e-2 & 5e-4 & 10 & 300 & - & SGD   & 0.7  & 0.5 & 0.4 & 0.5 & 0.4 \\
Pubmed    & 256 & 256 & 1e-3 & 5e-4 & 10 & 500 & -  & AdamW & 0.1  & 0.4 & 0.1 & 0.0 & 0.2      \\ 
DBLP      & 256 & 128 & 1e-3 & 5e-4 & 10 & 500 & - & AdamW & 0.8  & 0.1 & 0.2 & 0.2 & 0.3      \\ \hline
Flickr    & 512 & 128 & 1e-3 & 5e-4 & 20 & 200 & -  & AdamW & 0.1  & 0.0 & 0.2 & 0.2 & 0.2      \\
Reddit    & 512 & 128 & 1e-3 & 5e-4 & 20 & 200 & -  & AdamW & 0.2  & 0.4 & 0.1 & 0.0 & 0.2      \\ \hline
Cornell   & 64  & 256 & 1e-3 & 5e-4 & 10 & 200 & 20 & SGD   & 0.4  & 0.2 & 0.3 & 0.2 & 0.3      \\
Wisconsin & 64  & 256 & 1e-3 & 5e-4 & 10 & 200 & 20 & SGD   & 0.4  & 0.2 & 0.3 & 0.2 & 0.3      \\
Texas     & 64  & 256 & 1e-3 & 5e-4 & 10 & 200 & 20 & SGD   & 0.4  & 0.2 & 0.3 & 0.2 & 0.3      \\ \hline
CIAW      & 128 & 9   & 1e-2 & 5e-4 & -  & 200 & 20 & SGD   & 0.4  & 0.2 & 0.3 & 0.2 & 0.3      \\ \hline
\end{tabular}
\caption{Hyperparameters specifications}
\label{table:hyper}
\end{table*}

\subsection*{B. Dataset Details}
We will introduce the details of all datasets used in experiments. The statistics of datasets are in Table \ref{dataset}. All the datasets are available on Pytorch Geometry Library (PyG) \cite{fey2019fast}
\paragraph{Homophilous datasets}
We use four citation network datasets~\cite{sen2008collective}, Cora, Citeseer , Pubmed and DBLP. As for these datasets, nodes represent a variety of papers, and edges represent citation relationships between these papers. Node features are represented as the bag-of-word model of the corresponding paper, and the label is the academic topic of the paper. These four datasets are highly homophilous graphs that most of edges connect nodes sharing the same labels. Following GRACE, we also randomly split the nodes into $(10\%/10\%/80\%)$ for train/validation/test respectively instead of using the standard fixed splits which are unreliable for evaluating GNN methods~\cite{shchur2018pitfalls}. 

\paragraph{Non-homophilous datasets}
In real-world scenarios, the pattern "like attracts like" exists in many networks (\eg, friendship networks\cite{mcpherson2001birds}, citation networks~\cite{ciotti2016homophily}), but there also exists different pattern as "opposites attract" (\eg, dating networks or molecular networks~\cite{zhu2020homophily}). 
We can use edge homophily ratio $h=\frac{\left|\left\{(i, j):(i, j) \in \mathcal{E} \wedge y_{i}=y_{j}\right\}\right|}{|\mathcal{E}|}$ to indicate the portion of edges that connect two nodes with same labels.
A graph is considered to be non-homophilous if $h < 0.5$. 
% It is worth noting that non-homophilous and heterogeneous are different concepts\cite{sun2012mining}, where the heterogeneous graph means it consists of at least two kinds of nodes and various relationships between nodes. Exactly, homophilous and non-homophilous graphs are fine-grained types of homogeneous graphs.
Numerous GNN variants may fail on non-homophilous datasets because the aggregation functions can be considered as feature smoothing, and feature smoothing will average nodes’ features even if they have different labels. 
% In order to tackle this problem, some specifically designed GNNs\cite{pei2020geom},\cite{zhu2020homophily},\cite{zhu2020graph} emerge. 
% However, there are still few self-supervised graph learning methods that concentrate on such data, and pretext tasks of present works may not be suitable for such datasets. 
In order to verify that our method is more suitable for non-homophilous graphs, we also conduct experiments on three non-homophilous datasets collected by the CMU WebKB project, which are Cornell, Texas, and Wisconsin. These three datasets are webpage datasets collected from science departments of corresponding universities, where nodes represent web pages and edges represent hyperlinks between them. Node features are the bag-of-words representation of web pages, and these nodes are manually classified into five categories: Student, Project, Course, Staff, and Faculty. We use the preprocessed version in \cite{pei2020geom} with the standard dataset split. The detailed information of datasets is summarized in Table \ref{dataset}.

\paragraph{Inductive learning on large graphs}
We use two commonly used large graphs (Flickr and Reddit) to evaluate our method under inductive learning setting. Each node in Flickr represents an uploaded image. Edges are formed between nodes (images) from the same location, submitted to the same gallery, group, or set, images sharing common tags, images taken by friends, etc. Each node contains the 500-dimensional bag-of-word representation of the images provided by NUS-wide\footnote{https://lms.comp.nus.edu.sg/research/NUS-WIDE.htm}. And the labels are generated according to the tags of the images. Reddit is a social network with Reddit posts created in September 2014 which is preprocessed by \cite{hamilton2017inductive}. In the dataset, each node represents a post, and edges connect posts if the same user has commented on both. Each node contains 602-dimensional off-the-shelf GloVe word embeddings which are constructed from the post title, content, and comments, along with other metrics such as post score and the number of comments. We use the data splits processed by \cite{hamilton2017inductive}. Posts in the first 20 days are for training, including 151,708 nodes, and the remaining for testing (with $30\%$ data including 23,699 nodes for validation). The inductive setting follows \cite{velickovic2019deep}, validation and test nodes are invisible to the training algorithm.

\paragraph{Dynamic graph dataset}
We use one real-world dataset which is called The Contacts In A Workplace (CIAW)\footnote{http://www.sociopatterns.org/datasets/contacts-in-a-workplace/}. It is a vertex-focused dataset of [35] that contains the temporal network of contacts between individuals measured in an office building in France, from June 24, to July 3, 2013. Each node represents a worker wearing a sensor which can record the interaction with another worker within 1.5 m. The edges will be constructed between nodes if they have contacts (interaction lasting more than 20 s). For each 20s interval between June 24, and July 3, 2013, all the contacts occurring between the surveyed individuals (nodes) have been recorded. Each node is further characterized by his or her department name used as their labels. The task is to predict each individual’s department by leveraging the historical sequence of their interactions. For preprocessing CIAW, we downsample this dynamic graph into 20 discrete snapshots according to timestamp. The latest two snapshots are used for testing. Further, we randomly split nodes into 1:9 as training and testing nodes in each individual experiment. Moreover, we apply Node2vec \cite{10.1145/2939672.2939754} to generate 64-dimensional representation for each node by training graphs. For preprocessing CIAW*, we remove isolated nodes in each snapshot to simulate the situation of the addition and deletion of the nodes over time. For the training and testing process, in a semi-supervised manner, we will use training graphs to train the encoder through backwarding on labeled training nodes, and then test the performance of the encoder using testing nodes in testing graphs. In an unsupervised manner, we consider the adjacent snapshots as positive views because the evolution process is generally “smooth”. And the snapshots which are far away from the anchor are considered as negative views. Although we simply use static GNN encoder with discrete-time paradigms in our work, it can be applied to temporal GNN encoders with continuous-time paradigms. We leave this in future work.

\begin{table}[htb]
\centering
\begin{tabular}{cccccc}
\toprule
Dataset     & \# N     & \# E     & \# F         & \# C       & H \\ 
\midrule
Cora        & 2,078    & 5,278    & 1,433        & 7          & 0.81        \\
CiteSeer    & 3,327    & 4,676    & 3,703        & 6          & 0.74        \\
PubMed      & 19,717   & 44,327   & 500          & 3          & 0.80        \\
DBLP        & 17,716   & 105,734  & 1,639        & 4          & 0.83        \\ \hline
Cornell     & 183      & 280      & 1,703        & 5          & 0.30        \\
Texas       & 183      & 295      & 1,703        & 5          & 0.11        \\
Wisconsin   & 251      & 466      & 1,703        & 5          & 0.21        \\ \hline
Flickr      & 89,250   & 899,756  & 500          & 7          & 0.32        \\ 
Reddit      & 231,443  & 11,606,919  & 602       & 41         & 0.76        \\\hline
CIAW        & 92       & 9,827        & 64           & 5          & -           \\
\bottomrule
\end{tabular}
\caption{Details of used datasets, where we substitute N for \emph{Nodes}, E for \emph{Edges}, F for \emph{Features}, C for \emph{Classes}, H for \emph{Homophily ratio}.}
\label{dataset}
\end{table}

\subsection*{C. Implementation Details}
\paragraph{Model architecture}
We use two kinds of GNN encoders following \cite{zhu2020deep,velickovic2019deep}. On small-scale datasets, we adopt two layer GCN as:
\begin{align}
& \mathrm{GCN}_{i}(\mathbf{X}, \mathbf{A})=\sigma\left(\widehat{\mathbf{D}}^{-\frac{1}{2}} \widehat{\mathbf{A}} \widehat{\mathbf{D}}^{-\frac{1}{2}} \mathbf{X} \mathbf{W}_{i}\right), \\
& \mathbf{H} = \mathrm{GCN}_2(\mathrm{GCN}_1(\mathbf{X},\mathbf{A}),\mathbf{A}),
\end{align}
where $\hat{\mathbf{A}}=\mathbf{A}+\mathbf{I}$ is the adjacency matrix with self loop, and $\hat{\mathbf{D}}$ is the degree matrix of $\hat{\mathbf{A}}$. $\sigma$ is an activation function, $\mathbf{W}$ is a trainable linear transformation for input feature $\mathbf{X}$.

As for the large-scale datasets (Flickr and Reddit), we adopt a three-layer GraphSAGE-GCN~\cite{hamilton2017inductive} with residual connections as the encoder following DGI and GRACE:
\begin{align}
& \operatorname{MP}_{i}(\mathbf{X}, \mathbf{A})=\hat{\mathbf{D}}^{-1} \hat{\mathbf{A}} \mathbf{X} \mathbf{W}_{i} \\
& \widetilde{\operatorname{MP_i}}(\mathbf{X}, \mathbf{A})=\sigma\left(\mathbf{X} \mathbf{W}_{i}^{\prime} \| \operatorname{MP_i}(\mathbf{X}, \mathbf{A})\right) \\
& \mathbf{H}=\widetilde{\mathrm{MP}}_{3}\left(\widetilde{\mathrm{MP}}_{2}\left(\widetilde{\mathrm{MP}}_{1}(\mathbf{X}, \mathbf{A}), \mathbf{A}\right), \mathbf{A}\right).
\end{align}

\begin{figure*}[tb]
    \centering
    \includegraphics[scale=0.6]{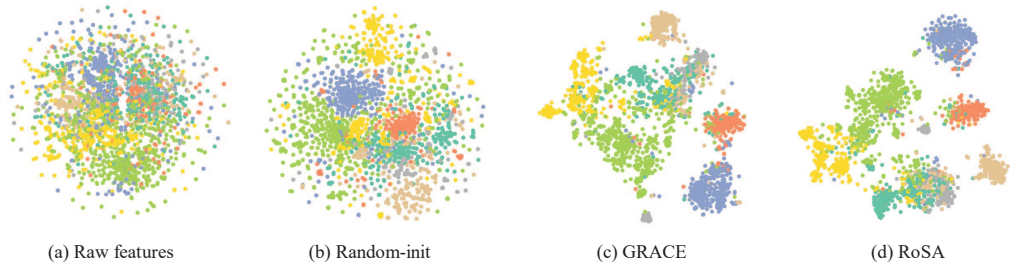}
    \caption{t-SNE visualization of node embeddings on Cora dataset, (a) is the raw features, (b) depicts features from a randomly initialized \our model, (c) shows embeddings from trained GRACE model, (d) is the result of trained \our. The margins of each cluster learned from \our are much wider than the learned GRACE.}
    \label{fig:vis}
\end{figure*}

\paragraph{Evaluation metrics}
We evaluate the learned encoder as follows. Firstly, we train the model in an unsupervised manner. Then, we extract node embeddings using the fixed pre-trained model. Lastly, a linear classifier will be trained on these embeddings across the training set and give the results on the test nodes. For four citation networks, we use an $l_2$-regularization LogisticRegression classifier from Scikit-Learn~\cite{pedregosa2011scikit} using the ‘liblinear’ solver following \cite{zhu2020deep}. For other datasets, we use one layer MLP through 100 epochs with Adam optimizer. We train the model for 20 runs and report the average classification accuracy or micro-averaged F1 score (on Flickr and Reddit) along with its standard deviation.

\paragraph{Computer infrastructures specifications}
For hardwares, all experiments are conducted on a computer server with eight GeForce RTX 3090 GPUs with 24GB memory and 64 AMD EPYC 7302 CPUs. Besides, our models are implemented by Pytorch Geometric 1.7.0~\cite{fey2019fast} and Pytorch 1.8.1~\cite{paszke2019pytorch}. All the datasets used in our work are available in PyTorch Geometric libraries.

% \subsection{Subsampling}
% We use random walk as our subsampling method, different naive random walk, we hope the positive pairs could contains some common nodes which 

\paragraph{Hyperparameters}

All hyperparameters used in experiments are listed in Table \ref{table:hyper}. $p_{e,1}, p_{e,2}, p_{f,1}, p_{f,2}$ are the probability parameters that control the extent of data augmentations like GRACE \cite{zhu2020deep}. $p_{e,1}, p_{e,2}$ is used for controlling the ratio of dropping edges and $p_{f,1}, p_{f,2}$ decides what a fraction of feature dimensions will be masked. For subsampling, %we set the walk length of random walk restart on all small-scale datasets as 10 and longer length on Flickr and Reddit,
we set the restart ratio as 0.8 on Pubmed and 0.5 on others. For one epoch, we only generate subgraphs for partial central nodes (limited by batch size). All models are initialized with Glorot initialization \cite{glorot2010understanding}. During the training process, we use an early stopping strategy on the observed results of the training loss with specific patience. \\
As for unsupervised adversarial training, we adopt that the inner loop runs 3 times ($M=3$) and set step size $\alpha$ as $10^{-3}$ to implicitly control perturbation budget $\epsilon$. The perturbation is not bounded by a definite $\epsilon$. The accumulated gradients for model parameters ($\theta, \omega$) during the inner loop will be used in the outer update like \cite{zhu2020freelb,kong2020flag}. In addition, we only add the adversarial perturbation $\delta$ to one view rather than two augmented views. \\
Concerning the order of augmentations, we firstly use subsampling to obtain a number of subgraphs, then edge dropping and feature masking will be applied on subgraphs. Lastly, an adversarial perturbation will be added to node features to improve model robustness. \\
Regarding sinkhorn algorithm \cite{cuturi2013sinkhorn}, we set the iteration number as 5 for computing transportation matrix $\mathbf{P}$ with $\lambda$ equaling to 20 in the regularization term. We find that tuning these two hyperparameters slightly changes performance because the main aim of energy transmutation is not changed.

\subsection*{D. Additional Experiments}
% \TODO{pass}

% \section{different loss}
% \begin{table}[]
% \begin{tabular}{lllll}
%     &            & citeseer & pubmed & cornell  \\
% n-g &            & 72.4±1.1 &        & 57.7±4.2 \\
% n-n & 83.9 ± 0.9 &          &        &          \\
% g-g &            &          &        &         
% \end{tabular}
% \end{table}

% using negative sampling

\paragraph{Sensitivity analysis}
% walk length
\begin{figure}[tb]
    \centering
    \includegraphics[scale=0.32]{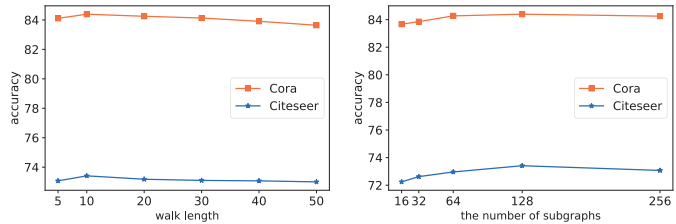}
    \caption{Analysis on critical hyperparameters. The left figure shows the impact of the walk length, the right figure embodies the influence of subgraph number.}
    \label{fig:walk_length}
\end{figure}
Firstly, we explore the influence of different walk length (steps) in sampling process. We measure how the performance is affected by varying walk length in the range of \{5, 10, 20, 30, 40, 50\}.
The results on Cora and Citeseer dataset are depicted in Figure \ref{fig:walk_length}. We get comparable results as the walk length reaches 10. After that, as the walk length gets larger, the accuracy drops. We guess that is because larger walk length (bigger subgraph) will introduce more noises. For instance, the negative samples are prone to contain more consistent substructures that can be considered positive signals, thus confusing the model to distinguish between positive and negative samples.

Secondly, we test the impact of different number of subgraphs trained in each epoch. This factor will determine the amount of negative samples during training. When the number reaches around 64, the accuracy becomes stable.

\subsection*{E. Visualization}
\begin{figure}[htb]
    \centering
    \includegraphics[scale=0.5]{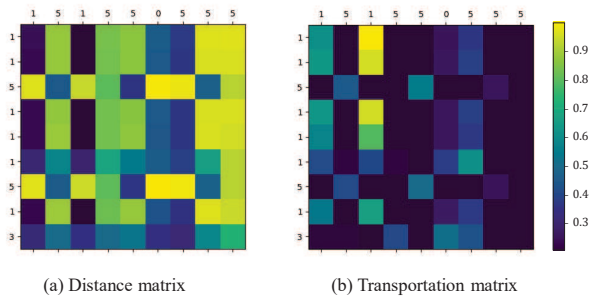}
    \caption{The distance matrix $\text{D}$ and transportation matrix $\Gamma$ of two contrasted views, where grid with high brightness has greater value. The x-axis means the labels of nodes in the first view and y-axis means labels in the second view. The energy transfer mostly occurs between nodes with the same category.}
    \label{fig:matrix}
\end{figure}

\paragraph{The visualization of one instance of EMD}
We visualize one contrasted pair of Cora in Figure \ref{fig:matrix}. As we can see, with two non-aligned subgraphs, introducing EMD can lead to a pseudo alignment process. More specifically, the distribution transport tends to happen more frequently between nodes with similar semantic meaning, which helps the model learn meaningful representations. 
% emd-matrix-vis

\paragraph{Embedding visualization}
In order to assess the quality of learned embeddings, we adopt t-SNE~\cite{van2008visualizing} to visualize the node embedding on Cora dataset using raw features, random-init of RoSA, GRACE, and RoSA , where different classes have different colors in Figure \ref{fig:vis}. We can observe that the 2D projection of node embeddings learned by RoSA has a clear separation of clusters, which indicates the model can help learn representative features for downstream tasks.

% \subsection*{E. Comparsion on More Datasets}

\end{document}